\pdfoutput=1

\documentclass[11pt]{article}

\usepackage{microtype}
\usepackage{xcolor}
\usepackage{graphicx}
\usepackage{comment}
\usepackage{array}
\usepackage{wrapfig,booktabs}
\usepackage{amssymb}
\usepackage{balance}
\usepackage{makecell}
\usepackage{tabularx, times}
\usepackage{latexsym}
\usepackage{authblk}

\usepackage[left=3.00cm, right=3.00cm, top=2.00cm, bottom=2.00cm]{geometry}
\usepackage{authblk}

\usepackage{naacl2021}

\graphicspath{{figures/}}

\usepackage[T1]{fontenc}

\usepackage[utf8]{inputenc}

\usepackage{microtype}

\title{They, Them, Theirs: Rewriting with Gender-Neutral English}

\author[$\dagger$]{Tony Sun}
\author[$\S$]{Kellie Webster}
\author[$\S$]{Apu Shah}
\author[$\dagger$]{William Yang Wang}
\author[$\S$]{Melvin Johnson}
\affil[$\dagger$]{University of California, Santa Barbara} 
\affil[$\S$]{Google Research}
\affil[ ]{\texttt {tonysun@ucsb.edu, william@cs.ucsb.edu}}
\affil[ ]{\texttt {\{websterk,apurva,melvinp\}@google.com}}
\date{}

\begin{document}
\maketitle

\begin{abstract}


Responsible development of technology involves applications being inclusive of the diverse set of users they hope to support.
An important part of this is understanding the many ways to refer to a person and being able to fluently change between the different forms as needed.
We perform a case study on the singular \emph{they}, a common way to promote gender inclusion in English.
We define a re-writing task, create an evaluation benchmark, and show how a model can be trained to produce gender-neutral English with $<$1\% word error rate with no human-labeled data.
We discuss the practical applications and ethical considerations of the task, providing direction for future work into inclusive natural language systems. 

\end{abstract}

\section{Introduction}

As the adoption of natural language models becomes more widespread, it is increasingly important to consider how they are used and who their users are.
One direction for considering this has been to ask whether representations encode social biases \cite{bolukbasi2018deb, gonen2019lipstick} or whether there is the potential for harm when applied to downstream applications \cite{blodgett2020language}, including coreference resolution \cite{zhao2018gender, rudinger2018gender, webster2018mind} and language generation \cite{sheng2019woman, nadeem2020stereoset}.
A complementary direction analyzes the distribution of training corpora and the potential to reinforce historical biases \cite{crawford2018trouble, Sun_2019}.

While much of this prior work studies gender identity, most is built on techniques which assume gender is binary.
At the same time, there is growing recognition of non-binary gender identities, with the singular \emph{they} emerging as a popular way to associate with a non-binary identity or simply not indicate a binary gender \citep{personalpronouns, americanpsychologicalassociation}.

Given its importance for gender inclusion, we perform a case study on the generation of the singular \emph{they} in NLP.
We define a new task, where we ask a model to understand and fluently change from gendered forms to neutral forms.
This captures alternations such as the he-pronoun sentence ``His dream is to be a fireman when he grows up'' becoming the they-pronoun counterpart: ``Their dream is to be a firefighter when they grow up''   (Figure~\ref{fig:rewriting_example}).
We do not consider sentences with more than one human entity, to avoid introducing ambiguity.
 
\begin{figure}[!tb]
\centering
\scalebox{1}
{\includegraphics[width=1\linewidth]{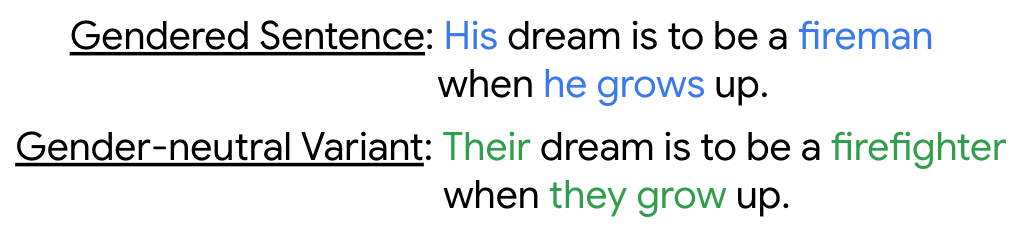}}
\caption{Example of rewriting a gendered sentence in English with its gender-neutral variant. Some challenges include resolving ``His'' as either ``Their'' or ``Theirs'' and rewriting the verb ``grows'' to be ``grow'' while keeping the verb ``is'' the same.} 
\label{fig:rewriting_example}
\end{figure}
 
We make two key contributions:
(1) we build a human-curated evaluation benchmark for the re-writing task, comprising 500 gendered and gender-neutral sentence-pairs from five domains (Twitter, Reddit, news articles, movie quotes, jokes); and
(2) we show how to train a model to produce they-forms in English \emph{with no human-labeled training data}.
We select lightweight, scalable techniques to allow future work to consider the many different identity markers across languages. 

The underlying goal of this work is to improve support for model users to be able to self-identify their preferred pronouns, thereby empowering people to interact with technology in a way that is consistent with their social identity.
We see applications of the task we propose in machine translation and augmented writing, and conclude by considering the ethical aspects of real-world deployment.


\section{Applications}

We see many applications of a model which can understand different reference forms and fluently move between them.
An obvious application is machine translation (MT), when translating from a language with gender-neutral pronouns (e.g. Turkish) to a language with gendered pronouns (e.g. English).
In this setting, many definitions of the MT task require a system to decide whether to use a he-pronoun or a she-pronoun in the English translation, even though this is not expressed in the source. If there is no disambiguating information available, providing multiple alternatives \cite{kuczmarski2018gender} with a gender-neutral translation might be more appropriate.
A model trained for this task could also be useful for augmented writing tools which provide suggestions to authors as they write, e.g.~to reduce unintended targeting in job listings.
Finally, given the efficacy of data augmentation techniques for improving model robustness \cite{dixon2018measuring}, we anticipate the sentences our re-writer produces would be useful input for models. 

\section{Challenges}

\noindent \textbf{Rule-based Approach.} One idea to rewrite gendered sentences in English to its gender-neutral variant is to use a simple rule-based method, mapping certain gendered words to their gender-neutral variants. This is a tempting solution, since the pronouns ``she'' and ``he'' can always be rewritten as ``they.'' In fact, we can even extend this idea to words such as ``fireman'' and ``spokesman,'' which can be mapped to their gender-neutral counterparts ``firefighter'' and ``spokesperson,'' respectively.

However, the simple find-and-replace approach struggles with two types of situations: (1) pronouns with one-to-many mappings and (2) identifying and rewriting verbs. In the former situation, some pronouns can be rewritten differently depending on the context of the sentence. To give an example, the pronoun ``her'' can map to either ``their'' or ``them'' depending on its usage. In the sentence ``This is \textit{her} pen,'' ``her'' should be rewritten as ``their,'' but in the sentence ``This pen belongs to \textit{her},'' ``her'' should be rewritten as ``them.'' Other examples of this one-to-many mapping include ``his'' to ``their / theirs'' and ``(s)he's'' to ``they're / they've.'' 

Another challenge that the rule-based method needs to address is how to identify and rewrite the appropriate verbs. From the example in Figure \ref{fig:rewriting_example}, the verb ``is'' remains the same while the verb ``grows'' is rewritten as ``grow.'' The insight is that verbs that correspond to a non-gendered subject should remain the same and verbs that correspond to either ``she'' or ``he'' as the subject should be rewritten. Doing this with a list of rules might be feasible, but would likely require a large set of handcrafted features that could be expensive to create and maintain.

\vspace{2mm}

\noindent \textbf{Seq2Seq Model.} On the other end of the spectrum, a separate idea is to train a Seq2Seq model that learns the mapping from gendered sentence to gender-neutral variant. This is a natural approach to the problem because the task of rewriting can be viewed a form of MT, a task where Seq2Seq models have demonstrated tremendous success \cite{sutskever2014sequence, bahdanau2014neural, wu2016google, vaswani2017attention}.

However, the main challenge here is that there are few naturally-occurring corpora for gendered and gender-neutral sentences. The lack of such a large, parallel dataset is a non-trivial challenge for the training and evaluation of a Seq2Seq model.

\begin{table*}[hbtp!]
\centering
\small
\footnotesize
\begin{tabularx}{\linewidth}{c c >{\centering\arraybackslash}X >{\centering\arraybackslash}X >{\centering\arraybackslash}X}
 \toprule
AC & MC & Original (gendered)  &  Algorithm & Model   \\ 
 \midrule
\checkmark & \checkmark & \textbf{Does she} know what happened to \textbf{her} friend? & \textbf{Do they} know what happened to \textbf{their} friend? & \textbf{Do they} know what happened to \textbf{their} friend?  \\ 
\checkmark & & Manchester United boss admits failure to make top four could cost \textbf{him his} job & Manchester United boss admits failure to make top four could cost \textbf{them their} job  & Manchester United boss admits failure to make top four could cost \textbf{them theirjob} \\
& \checkmark & \textbf{She sings} in the shower and \textbf{dances} in the dark. & \textbf{They sing} in the shower and \textbf{dances} in the dark. & \textbf{They sing} in the shower and \textbf{dance} in the dark.   \\ 
 \bottomrule
\end{tabularx}
\caption{Comparison of algorithm and model on different examples. Algorithm Correct (AC) and Model Correct (MC) indicates that whether the algorithm or model is correct. In the second example, the model unnecessarily removes a whitespace. In the third example, the algorithm does not pluralize the verb ``dances'' due to a mistake with the underlying dependency parser.}
\label{gn_examples}
\end{table*}







\begin{figure}[!tb]
\centering
\scalebox{1}
{\includegraphics[width=1\linewidth]{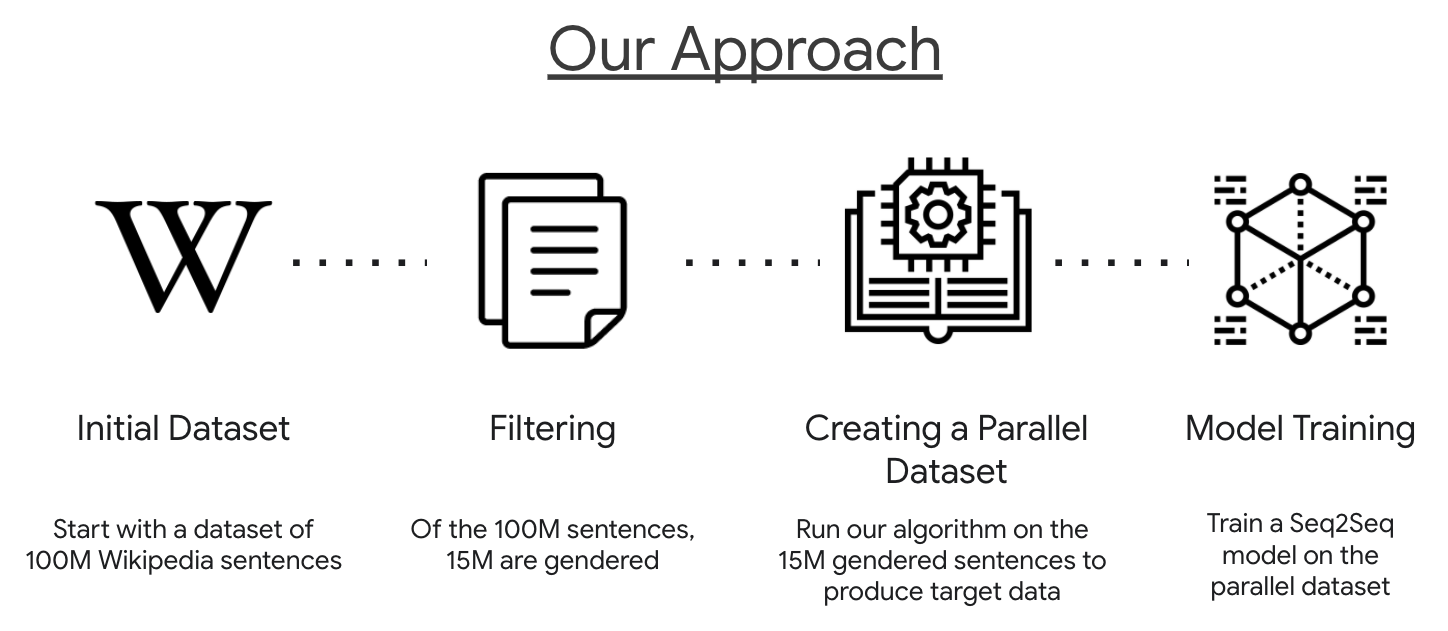}}
\caption{Our approach for rewriting without using a pre-existing parallel dataset. We start with an initial corpora from Wikipedia, filter and keep the gendered sentences, create a rewriting algorithm to create a parallel dataset, and train a Seq2Seq model on that dataset.} 
\label{fig:rewriting_example}
\end{figure}

\section{Methodology}

Our approach aims to combine the positive aspects of both the rule-based approach and Seq2Seq model. Although parallel gendered and gender-neutral sentences can be difficult to mine, standalone gendered sentences are easily accessible. We take a corpora of 100 million sentences from Wikipedia and filter them using regular expressions based on our criteria of containing a single gendered entity. \footnote{We would keep sentences that contain words from the set $\{she, her, hers, herself\}$ or the set $\{he, him, his, himself\}$, but not sentences that contain words from both sets since that would indicate multiple entities of different genders.} Using these filters, we find that 15 million sentences from our original Wikipedia corpora are gendered and take them to be the source side of our dataset.
\renewcommand{\arraystretch}{1.5}

From these gendered sentences, we form a parallel dataset by creating a rewriting algorithm that takes a gendered sentence as input and returns its gender-neutral variant as output. Then, we train a Seq2Seq model on this parallel dataset to learn the mapping from gendered to gender-neutral sentences. Building an end-to-end Seq2Seq model enables us to: 1) generalize better to unseen examples, and 2) convert the multi-step algorithm into a single step.

Finally, we evaluate the performance of both the algorithm and the model on a test set of 500 manually annotated sentence-pairs \footnote{We plan to open-source the code for the algorithm and the resulting  model.}. Source sentences, which are gendered, in the test set are taken equally from five diverse domains: Twitter, Reddit, news articles, movie quotes, and jokes. In addition, we ensure that each domain is gender-balanced, meaning that we use an equal number of sentences with masculine and feminine pronouns from each domain. It is important to have a gender-balanced test set so that a rewriter cannot attain a strong score by performing well on either masculine or feminine sentences. For our metrics, we use BLEU score \cite{papineni2002bleu} and word error rate (WER). Comparisons between the algorithm and the model are shown in Table \ref{gn_examples}. We elaborate on both the algorithm and the model below in further detail.

\subsection{Rewriting Algorithm}
Our rewriting algorithm is composed of three main components: regular expressions, a dependency parser, and a language model. 

Regular expressions are responsible for finding and replacing certain tokens regardless of the context. Similar to the aforementioned rule-based approach, we always rewrite ``(s)he'' to be ``they'' and certain stereotypically gendered words to their gender-neutral counterparts. \footnote{We release full list of words in the Appendix.}

We use SpaCy's~\cite{spacy2} dependency parser for building a parse tree of the input sentence. Using the parse tree, we tag verbs that correspond to ``(s)he'' as their subject and convert them to  their third-person plural conjugation using a list of rules. Verbs that correspond to a non-gendered subject remain the same.

Finally, we use the pre-trained GPT-2 language model~\cite{radford2019language} to resolve pronouns with one-to-many mappings. Given a sentence with such a pronoun, we would generate all possible variants by substituting each gender-neutral variant. Then, we rank the perplexity of each sentence and choose the sentence with the lowest perplexity as the gender-neutral variant.




\subsection{Model Details}
We train a Transformer model \cite{vaswani2017attention} with $6$ encoder and $6$ decoder layers using the Fairseq library \cite{Ott_2019} on the parallel dataset generated by our algorithm. We further augment our parallel data with: 1) non-gendered to non-gendered identity data and 2) gender-inflected sentences (converting a masculine sentence to a feminine sentence or vice versa and keeping the same gender-neutral translation). We find that both of these augmentation techniques lead to improved performance. We use a split of 70-30 rewritten and identity pairs in the training set.

Our $d_{model}$ and $d_{ff}$ values are 512 and 1024 respectively and we use 4 attention heads. We use Adam~\cite{kingma2014adam} with an initial learning rate of $5e-4$ and the inverse square-root schedule described in~\cite{vaswani2017attention} with a warmup of $4000$ steps. Additionally, we use a byte-pair vocabulary~\cite{sennrich2015neural} of $32000$ units. Finally, we pick the checkpoint with the best BLEU score on the dev set (held-out from Wikipedia).

\begin{table}[t]
\centering
\small
\footnotesize
\begin{tabular}{l  c  c  c} 
 \toprule
            & BLEU  & WER  \\ 
 \midrule
 Source (identity)    & 90.32 & 12.40\%  \\ 
 Algorithm  & \textbf{99.63} & \textbf{0.63\%} \\
 Model      & 99.44 & 0.99\%   \\ 
 \bottomrule
\end{tabular}
\caption{Metrics on the test set}
\label{gn_results}
\end{table}

\section{Results}

We present our results on the annotated test set in Table \ref{gn_results}. We include the source sentence to establish the baseline with the identity function, which demonstrates that only certain gendered terms and their corresponding verbs should be rewritten while all other tokens remain the same.

\begin{figure}[t]
\centering
\scalebox{1}
{\includegraphics[width=1\linewidth]{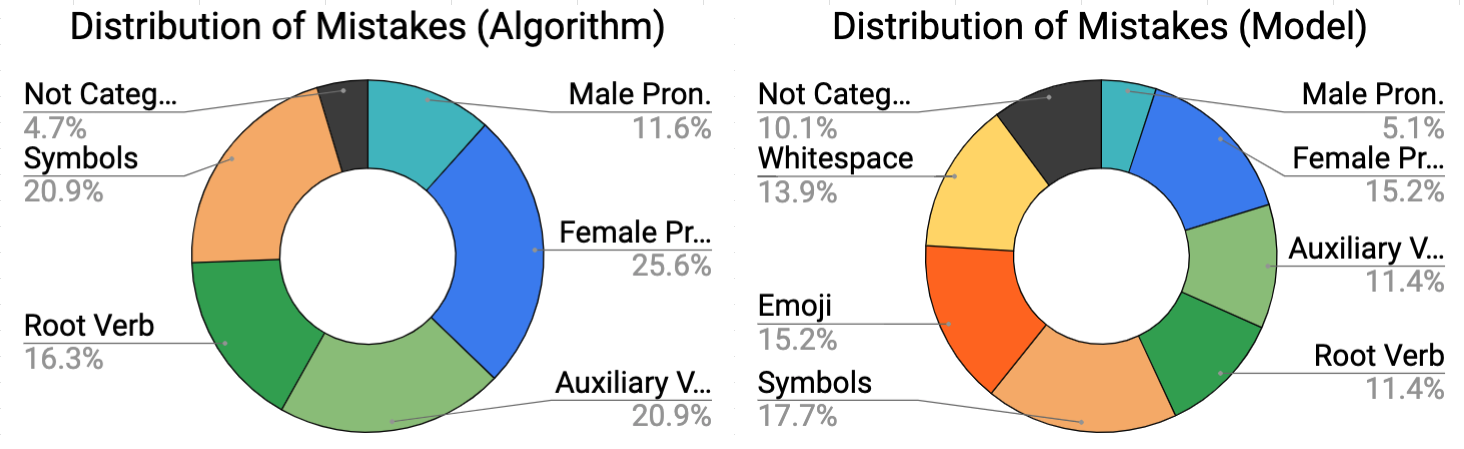}}
\caption{Distribution of mistakes of algorithm compared to model. Nearly three-quarters of the mistakes the algorithm makes are due to pronouns and verb, whereas less than half of the model's mistakes are due those categories. The} 
\label{fig:mistake_distribution}
\end{figure}

Impressively, both the algorithm and the model achieve over 99 BLEU and less than 1\% word error rate. We find that the algorithm performs marginally better than the model on the test set. However, upon closer inspection, as seen in Figure \ref{fig:mistake_distribution}, the model actually has less of its mistakes due to pronouns and verbs, but nearly half of the model's mistakes are due to rare tokens like whitespaces, emojis, and symbols (e.g. @, \%, *, etc.). The algorithm will occasionally make mistakes when the underlying dependency parser incorrectly categorizes the part of speech of a target verb, as seen in the third example in Table \ref{gn_examples}, or when the language model gives the wrong gender-neutral variant a lower perplexity.

Robustness to rare words and domain mismatch are known problems for Seq2Seq models~\cite{koehn2017six}. In this situation, this problem may likely be exacerbated by the domain mismatch between training (clean Wikipedia sentences) and testing (noisy sentences from Twitter and Reddit). In fact, the model performs better on domains such as movie quotes and jokes, which tend to contain clean text.

We also find that both the algorithm and model tend to make more mistakes on feminine sentences compared to masculine sentences. For the algorithm, we noticed that the language model can have difficulty resolving ``her'' to ``their / them'' in certain contexts. For the model, we observed that around 70\% of gendered sentences from Wikipedia were masculine. We aimed to offset this with the gender-inflected sentences, but having more naturally-occurring feminine sentences would likely help improve model performance.





\section {Conclusion}

We have presented a case study into how to fluently move between different ways to refer to a person, focusing on the alternation between English gendered pronouns and the singular \emph{they}.
We propose the new task of re-writing gendered sentences to be gender-neutral with the underlying goal of developing gender inclusive natural language systems. 
We show how it is possible to train a model to do this without any human-labeled data, by automatically generating large amounts of data by rule.

It is important to recognize that there are very many different identity markers in the many languages beyond English, and we are excited by the potential for our lightweight methodology to scale to these.
We are excited for future work to dig into challenges with these languages, especially those which have greater morphological complexity.

\section{Ethical Considerations}

An underlying aim of the task we propose is to improve support for model users to be able to self-identify their preferred pronouns, including if these are non-binary.
That is, the task follows and reflects the social change in English of using the singular they for gender inclusion.
Work in this direction should always aim to be empowering, to allow people to interact with technology in a way that is consistent with their social identity.
Therefore, an anti-goal of this work that should be avoided in future work, is being able to infer a person's preferred pronouns when they are not explicitly given to a system.
It is also important to note that we do not prescribe that a particular form is correct or preferred to others, only that it is a desirable capability for models to be able to understand the different forms and fluently change between them as needed.
As alternative forms emerge, such as \emph{ze}, it is important that language models adapt to understand and be able to produce these also. For increased transparency, we provide a detailed model card~\cite{mitchell2019model} in the Appendix.

\bibliography{anthology,custom}

\begin{thebibliography}{26}
\expandafter\ifx\csname natexlab\endcsname\relax\def\natexlab#1{#1}\fi

\bibitem[{Bahdanau et~al.(2014)Bahdanau, Cho, and Bengio}]{bahdanau2014neural}
Dzmitry Bahdanau, Kyunghyun Cho, and Yoshua Bengio. 2014.
\newblock \href {http://arxiv.org/abs/1409.0473} {Neural machine translation by
  jointly learning to align and translate}.

\bibitem[{Blodgett et~al.(2020)Blodgett, Barocas, au2, and
  Wallach}]{blodgett2020language}
Su~Lin Blodgett, Solon Barocas, Hal Daumé~III au2, and Hanna Wallach. 2020.
\newblock \href {http://arxiv.org/abs/2005.14050} {Language (technology) is
  power: A critical survey of "bias" in nlp}.

\bibitem[{Bolukbasi et~al.(2018)Bolukbasi, Chang, Zou, Saligrama, and
  Kalai}]{bolukbasi2018deb}
Tolga Bolukbasi, Kai-Wei Chang, James~Y Zou, Venkatesh Saligrama, and Adam~T
  Kalai. 2018.
\newblock Debiaswe.
\newblock \url{https://bit.ly/2ruopBZ}.
\newblock Accessed on 12.10.2018.

\bibitem[{Crawford(2017)}]{crawford2018trouble}
Kate Crawford. 2017.
\newblock {The Trouble With Bias}.
\newblock Keynote at Neural Information Processing Systems (NIPS'17).

\bibitem[{Dixon et~al.(2018)Dixon, Li, Sorensen, Thain, and
  Vasserman}]{dixon2018measuring}
Lucas Dixon, John Li, Jeffrey Sorensen, Nithum Thain, and Lucy Vasserman. 2018.
\newblock Measuring and mitigating unintended bias in text classification.
\newblock In \emph{Proceedings of the 2018 AAAI/ACM Conference on AI, Ethics,
  and Society}, pages 67--73. ACM.

\bibitem[{Gonen and Goldberg(2019)}]{gonen2019lipstick}
Hila Gonen and Yoav Goldberg. 2019.
\newblock {Lipstick on a Pig: Debiasing Methods Cover up Systematic Gender
  Biases in Word Embeddings But Do Not Remove Them}.
\newblock In \emph{North American Chapter of the Association for Computational
  Linguistics (NAACL'19)}.

\bibitem[{Honnibal and Montani(2017)}]{spacy2}
Matthew Honnibal and Ines Montani. 2017.
\newblock {spaCy 2}: Natural language understanding with {B}loom embeddings,
  convolutional neural networks and incremental parsing.
\newblock To appear.

\bibitem[{Kingma and Ba(2014)}]{kingma2014adam}
Diederik~P Kingma and Jimmy Ba. 2014.
\newblock Adam: A method for stochastic optimization.
\newblock \emph{arXiv preprint arXiv:1412.6980}.

\bibitem[{Koehn and Knowles(2017)}]{koehn2017six}
Philipp Koehn and Rebecca Knowles. 2017.
\newblock Six challenges for neural machine translation.
\newblock \emph{arXiv preprint arXiv:1706.03872}.

\bibitem[{Kuczmarski and Johnson(2018)}]{kuczmarski2018gender}
James Kuczmarski and Melvin Johnson. 2018.
\newblock Gender-aware natural language translation.

\bibitem[{Lee(2019)}]{americanpsychologicalassociation}
Chelsea Lee. 2019.
\newblock \href {https://apastyle.apa.org/blog/singular-they} {Welcome,
  singular "they"}.

\bibitem[{Mitchell et~al.(2019)Mitchell, Wu, Zaldivar, Barnes, Vasserman,
  Hutchinson, Spitzer, Raji, and Gebru}]{mitchell2019model}
Margaret Mitchell, Simone Wu, Andrew Zaldivar, Parker Barnes, Lucy Vasserman,
  Ben Hutchinson, Elena Spitzer, Inioluwa~Deborah Raji, and Timnit Gebru. 2019.
\newblock Model cards for model reporting.
\newblock In \emph{Proceedings of the conference on fairness, accountability,
  and transparency}, pages 220--229.

\bibitem[{Nadeem et~al.(2020)Nadeem, Bethke, and Reddy}]{nadeem2020stereoset}
Moin Nadeem, Anna Bethke, and Siva Reddy. 2020.
\newblock \href {http://arxiv.org/abs/2004.09456} {Stereoset: Measuring
  stereotypical bias in pretrained language models}.

\bibitem[{Ott et~al.(2019)Ott, Edunov, Baevski, Fan, Gross, Ng, Grangier, and
  Auli}]{Ott_2019}
Myle Ott, Sergey Edunov, Alexei Baevski, Angela Fan, Sam Gross, Nathan Ng,
  David Grangier, and Michael Auli. 2019.
\newblock \href {https://doi.org/10.18653/v1/n19-4009} {fairseq: A fast,
  extensible toolkit for sequence modeling}.
\newblock \emph{Proceedings of the 2019 Conference of the North}.

\bibitem[{Papineni et~al.(2002)Papineni, Roukos, Ward, and
  Zhu}]{papineni2002bleu}
Kishore Papineni, Salim Roukos, Todd Ward, and Wei-Jing Zhu. 2002.
\newblock Bleu: a method for automatic evaluation of machine translation.
\newblock In \emph{Proceedings of the 40th annual meeting of the Association
  for Computational Linguistics}, pages 311--318.

\bibitem[{Radford et~al.(2019)Radford, Wu, Child, Luan, Amodei, and
  Sutskever}]{radford2019language}
Alec Radford, Jeff Wu, Rewon Child, David Luan, Dario Amodei, and Ilya
  Sutskever. 2019.
\newblock Language models are unsupervised multitask learners.

\bibitem[{Rudinger et~al.(2018)Rudinger, Naradowsky, Leonard, and
  Van~Durme}]{rudinger2018gender}
Rachel Rudinger, Jason Naradowsky, Brian Leonard, and Benjamin Van~Durme. 2018.
\newblock {Gender Bias in Coreference Resolution}.
\newblock In \emph{North American Chapter of the Association for Computational
  Linguistics (NAACL`18)}.

\bibitem[{Sakurai(2017)}]{personalpronouns}
Shige Sakurai. 2017.
\newblock \href {https://www.mypronouns.org/what-and-why} {Personal pronouns,
  what and why}.

\bibitem[{Sennrich et~al.(2015)Sennrich, Haddow, and
  Birch}]{sennrich2015neural}
Rico Sennrich, Barry Haddow, and Alexandra Birch. 2015.
\newblock Neural machine translation of rare words with subword units.
\newblock \emph{arXiv preprint arXiv:1508.07909}.

\bibitem[{Sheng et~al.(2019)Sheng, Chang, Natarajan, and Peng}]{sheng2019woman}
Emily Sheng, Kai-Wei Chang, Premkumar Natarajan, and Nanyun Peng. 2019.
\newblock \href {http://arxiv.org/abs/1909.01326} {The woman worked as a
  babysitter: On biases in language generation}.

\bibitem[{Sun et~al.(2019)Sun, Gaut, Tang, Huang, ElSherief, Zhao, Mirza,
  Belding, Chang, and Wang}]{Sun_2019}
Tony Sun, Andrew Gaut, Shirlyn Tang, Yuxin Huang, Mai ElSherief, Jieyu Zhao,
  Diba Mirza, Elizabeth Belding, Kai-Wei Chang, and William~Yang Wang. 2019.
\newblock \href {https://doi.org/10.18653/v1/p19-1159} {Mitigating gender bias
  in natural language processing: Literature review}.
\newblock \emph{Proceedings of the 57th Annual Meeting of the Association for
  Computational Linguistics}.

\bibitem[{Sutskever et~al.(2014)Sutskever, Vinyals, and
  Le}]{sutskever2014sequence}
Ilya Sutskever, Oriol Vinyals, and Quoc~V Le. 2014.
\newblock Sequence to sequence learning with neural networks.
\newblock In \emph{Advances in neural information processing systems}, pages
  3104--3112.

\bibitem[{Vaswani et~al.(2017)Vaswani, Shazeer, Parmar, Uszkoreit, Jones,
  Gomez, Kaiser, and Polosukhin}]{vaswani2017attention}
Ashish Vaswani, Noam Shazeer, Niki Parmar, Jakob Uszkoreit, Llion Jones,
  Aidan~N Gomez, {\L}ukasz Kaiser, and Illia Polosukhin. 2017.
\newblock Attention is all you need.
\newblock In \emph{Advances in neural information processing systems}, pages
  5998--6008.

\bibitem[{Webster et~al.(2018)Webster, Recasens, Axelrod, and
  Baldridge}]{webster2018mind}
Kellie Webster, Marta Recasens, Vera Axelrod, and Jason Baldridge. 2018.
\newblock {Mind the GAP: A Balanced Corpus of Gendered Ambiguous Pronouns}.
\newblock In \emph{Transactions of the ACL (TACL'18)}.

\bibitem[{Wu et~al.(2016)Wu, Schuster, Chen, Le, Norouzi, Macherey, Krikun,
  Cao, Gao, Macherey et~al.}]{wu2016google}
Yonghui Wu, Mike Schuster, Zhifeng Chen, Quoc~V Le, Mohammad Norouzi, Wolfgang
  Macherey, Maxim Krikun, Yuan Cao, Qin Gao, Klaus Macherey, et~al. 2016.
\newblock Google's neural machine translation system: Bridging the gap between
  human and machine translation.
\newblock \emph{arXiv preprint arXiv:1609.08144}.

\bibitem[{Zhao et~al.(2018)Zhao, Wang, Yatskar, Ordonez, and
  Chang}]{zhao2018gender}
Jieyu Zhao, Tianlu Wang, Mark Yatskar, Vicente Ordonez, and Kai-Wei Chang.
  2018.
\newblock {Gender Bias in Coreference Resolution: Evaluation and Debiasing
  Methods}.
\newblock In \emph{North American Chapter of the Association for Computational
  Linguistics (NAACL`18)}.

\end{thebibliography}
\bibliographystyle{acl_natbib}

\clearpage
\appendix

\section{Appendix}
\label{sec:appendix}

List of mappings from stereotypically gendered terms to their gender-neutral counterparts (this is by no means comprehensive, and there is an active area of research on expanding this list):

$\textrm{mankind} \rightarrow \textrm{humanity}$

$\textrm{layman} \rightarrow \textrm{layperson}$

$\textrm{laymen} \rightarrow \textrm{lay people}$

$\textrm{policeman} \rightarrow \textrm{police officer}$

$\textrm{policewoman} \rightarrow \textrm{police officer}$

$\textrm{policemen} \rightarrow \textrm{police officers}$

$\textrm{policewomen} \rightarrow \textrm{police officers}$

$\textrm{stewardess} \rightarrow \textrm{flight attendant}$

$\textrm{weatherman} \rightarrow \textrm{weather reporter}$

$\textrm{fireman} \rightarrow \textrm{firefighter}$

$\textrm{chairman} \rightarrow \textrm{chair}$

$\textrm{spokesman} \rightarrow \textrm{spokesperson}$

\vspace{5mm}

{\Large \textbf{Model Card}}

\textbf{Model Details}
\begin{itemize}
\item Developed by researchers at the University of California, Santa Barbara and Google.
\item Transformer model with 6 encoder layers and 6 decoder layers.
\item Trained to optimize per-token cross-entropy loss.
\end{itemize}

\textbf{Intended Use}
\begin{itemize}
\item Used for research in converting gendered English sentences with single person entities, into gender-neutral forms.
\item Not suitable for generic text that has not been preselected for having single individual entities.
\end{itemize}

\textbf{Factors}
\begin{itemize}
\item Based on training data and difficulty the results for converting he-pronoun sentences to neutral vs she-pronoun sentences to neutral vary as documented in Figure 3.
\end{itemize}

\textbf{Metrics}
\begin{itemize}
\item Evaluation metrics include BLEU and WER.
\end{itemize}

\textbf{Training Data}
\begin{itemize}
\item 15 million sentences filtered from 100 million English Wikipedia sentences containing single gendered entities. 
\item These sentences were automatically rewritten using manual rules plus the SpaCy dependency parser and GPT-2 LM. 
\end{itemize}

\textbf{Evaluation Data}
\begin{itemize}
\item Manually curated English gendered sentences represented equally across  five diverse domains: Twitter, Reddit, news articles, movie quotes, and jokes.
\end{itemize}

\textbf{Ethical Considerations}
\begin{itemize}
\item The technology is intended to be used to demonstrate the ability of models to rewrite gendered text to the chosen gender of a user. 
\item The model is not intended to identify gender nor to prescribe a particular form.
\item This is by no-means exhaustive and does not match all potential pronouns a user may request.
\end{itemize}

\textbf{Caveats}
\begin{itemize}
\item Intended to rewrite sentences only consisting of a single person entity.
\end{itemize}

\end{document}